\documentclass{article}
\usepackage{graphicx} 
\usepackage{amsmath, amssymb}
\usepackage{svg}
\usepackage{url}
\usepackage{algorithm}
\usepackage{algpseudocode}
\usepackage{listings}
\usepackage{xcolor}
\begin{document}

\begin{center}
{\Large\bfseries CacheRouter: A Dual-Path Tool Routing Architecture with Cache-Preserving Main-Model Isolation for Long-Tail Tool Discovery}\\[1.2em]
{\large Donghui Zha\textsuperscript{1}, Lingwei Xu\textsuperscript{1},\\
Linxiao Wu\textsuperscript{2}, Yixue Dong\textsuperscript{3}, Haochen Li\textsuperscript{1,*}}\\[0.6em]
{\normalsize \textsuperscript{1}School of Mathematical Sciences, Beijing University of Posts and Telecommunications, Beijing 100876, China\\
\textsuperscript{2}School of Mathematics and Statistics, Chongqing University, Chongqing 401331, China\\
\textsuperscript{3}School of Science, Beijing Forestry University, Beijing 100083, China}\\[0.6em]
{\small * Corresponding author. E-mail address: \texttt{lihaochen\_bjut@sina.com} (Haochen Li).}
\end{center}

\vspace{0.5em}

\begin{center}
\textbf{Abstract}
\end{center}

\noindent Tool use in LLM systems faces a structural trade-off. Progressive disclosure keeps the prompt small by showing only the tools relevant to the current task, while prompt caching rewards a request prefix that stays fixed across calls; every change to the visible tool list invalidates the cached prefix. This paper treats the trade-off as a problem of request architecture and proposes a dual-path routing design that assigns tool selection and tool delivery to separate channels. The main model always sees a small, fixed set of core tools, so the head of its request is unchanged across calls; all other tools are reached through an independent routing channel, in which a router sub-model searches the full tool list, selects one tool, executes it, and returns the result. Tool registration is automated from source code and supports runtime updates, so the tool set can grow without modifying the main model's request prefix. The design generalizes progressive disclosure: capabilities are disclosed through the routing channel, and the main model's prefix stays stable. A prototype implementation was exercised on 55 functional queries and a 30-turn dialogue; token-level cache hit rates reached 90.99\% and 95.2\%, cutting input cost to about 12.0\% and 8.0\% of a no-cache baseline under DeepSeek's pricing, where cache-hit input tokens cost roughly 1/30 of cache-miss tokens.


\noindent\textbf{Code:} \url{https://github.com/qingyiwanchen/CacheRouter}

\section{Introduction}

\subsection{Research Background and Motivation}

Large language models are increasingly used as agents that call external tools to complete tasks (Achiam et al. 2023; Dubey et al. 2024; Guo et al. 2025). Through function calling, a model can fetch real-time information, run code, read and write files, or query third-party APIs instead of relying only on its training data. The Model Context Protocol (MCP) (Anthropic 2026) standardizes how tools are described and invoked, so one agent runtime can load tools from different providers through a single interface. The runtime that binds a model to its tools (the tool-use harness) decides which tool definitions enter each request and when they change.

More tools mean more capabilities, but they also make every request larger: each tool contributes a name, a parameter schema, and a description, all of which are sent to the model as part of the prompt. With a handful of tools this overhead is negligible, but once the collection grows to hundreds or thousands, sending every tool definition with every request becomes costly in both tokens and latency. Tool definitions also affect caching, because they sit near the front of the request: changing them between two requests invalidates the prefix that the API cache reuses.

The cost is concrete: providers such as DeepSeek (DeepSeek-AI 2024; DeepSeek 2026) bill cache-hit input tokens at a small fraction of cache-miss tokens. In multi-turn agents, the system prompt, history, and tool definitions all sit inside the cached prefix, so changing the tool definitions between adjacent requests forfeits that discount even for nearly identical tasks.

Traditional orchestration binds the two decisions into one request: the model picks tools, and the application injects their definitions. That works while the tool set is small; as the set changes, the tool segment of the request changes with it, a coupling we summarize as ``the more flexible the tools, the less stable the cache.'' The opposite extreme, always sending all tools, keeps the prefix fixed but wastes tokens on irrelevant definitions.

This is a conflict between two goals. Progressive disclosure, which shows relevant tools step by step, keeps context small but changes the request prefix; prompt caching needs that prefix to stay stable. The two pull in opposite directions.

A cache hit means the same input prefix is reused, not that the same query returns the same answer. If disclosure changes that prefix, a nearly identical task still misses the cache, and the tokens saved by showing fewer tools are easily outweighed by re-processing the larger prefix later.

The central question is whether tool-set flexibility, prefix stability, and full tool coverage can be maintained at the same time while the tool set keeps evolving; retrieval accuracy is not the binding constraint.

The root cause is that two processes are merged: \textbf{tool selection} (what capabilities the task needs) and \textbf{tool delivery} (which definitions enter the request). In the same channel, every selection change rewrites the prefix, so tool-set dynamism propagates directly into the cache layer.

Our position is that the two need not share a request. Frequent, stable core tools stay in a fixed view with a stable prefix; infrequent or generated long-tail tools are discovered and run through a separate routing channel. The ecosystem can then grow without restructuring the main request path.

This paper develops this position into an architecture, \textbf{CacheRouter}, and implements a prototype of it. The name reflects the core idea: cache preservation is the constraint, dual-path routing is the organizing principle. The system uses MCP as its tool protocol, keeps a fixed core tool set as the main model's stable view, connects the main model to long-tail tools through \texttt{INTERNAL\_ROUTER}, and combines automatic registration, hot reloading, and skill bootstrapping into one loop spanning tool discovery, routing, and evolution.

\subsection{Problems and Challenges}

Orchestrating tools at scale raises three problems that existing work treats separately: how to register and maintain many tools, how to keep context and cache behavior under control as the tool set grows, and how to let the agent acquire new capabilities at runtime.

\subsubsection{Tool Registration and Engineering Scalability}

With a few tools, hand-maintaining their names, parameter schemas, and server-side registration logic is acceptable. As the set grows, this manual work becomes the bottleneck: every new tool means updating a schema and registration code by hand.

The fix is to separate tool implementation from tool registration: a developer drops a script into a directory, and the system parses its arguments, generates the schema, and registers it. Registration must also keep up with changes at runtime, without stopping and redeploying the service.

Registration therefore needs both automation and service continuity: the repository evolves while in-flight requests keep working, which constrains registry updates, index consistency, and concurrent access.

\subsubsection{Engineering Trade-off between Cache Efficiency and Tool Scale}

Let $M$ be the number of registered tools. When $M$ is a dozen or so, sending every tool definition with each request is fine: the payload is small and the request prefix stays fixed.

At hundreds or thousands of tools, full loading turns tool definitions into a large fixed payload: low-frequency tools are resent on every request even when irrelevant, inflating token counts without contributing to decisions.

Dynamic selection, sending only the tools relevant to the current query, shrinks a single request but makes the tool segment change between requests, which undermines cache reuse.

``Fewer tools'' is not the right objective by itself. The real target is a balance between \textbf{context size} and \textbf{prefix stability}: keep the prefixes of high-frequency requests stable while controlling the tool-definition payload.

We treat this as an architecture problem: one request should not have to search a large tool space and keep a stable prefix at the same time. The two jobs are split: the main model keeps a stable core set of controllable size, and an independent routing channel retrieves and invokes the rest. Tool growth then happens outside the main model's cache path.

\subsubsection{Autonomous Runtime Capability Expansion}

Open-ended agents constantly encounter new tasks beyond the coverage of the existing tool set in real environments. If a system can only use the fixed tools registered at deployment time, its capability boundary is also tied to the tool repository. For long-running agents, such a static capability boundary gradually becomes a limitation on sustained service ability.

Ideally, when the system finds that no existing tool can accomplish a task, it should be able to generate new tools through its existing code-execution or skill-construction capabilities, consolidate them into standard skills, and incorporate them into the tool repository. This forms a bootstrapping loop of ``capability-gap discovery, ad-hoc implementation, validation, skill consolidation, and later reuse.''

Bootstrapping also conflicts with caching: adding every generated skill to the main model's fixed tool set would change its tool definitions and break existing cache prefixes. Expansion must also answer ``how to add tools without changing the request path.''

We therefore place runtime-bootstrapped skills in an independent dynamic tool space, accessed on demand through routing, so generating a new tool never expands the main model's fixed view; capability grows while the core request path stays stable.

\subsection{Contributions}

To address the problems above, this paper presents \textbf{CacheRouter}, an architecture for tool orchestration that uses MCP as its unified tool protocol and treats cache stability as a first-class design constraint, together with a prototype implementation. The methodological core is a reconciliation of two constraints usually treated as opposites: progressive disclosure, which keeps context small by revealing tools gradually, and prompt caching, which needs the request prefix to stay fixed. The resolution is structural, moving disclosure out of the main request into an independent routing channel. The main contributions are summarized in the following five aspects.

\textbf{(1) A cache-aware dual-path tool execution architecture.} This paper decouples ``tool selection'' from ``tool delivery'' at the system-architecture level, partitioning the tool space into a core tool set that is fixed and visible to the main model and a dynamic long-tail tool set managed by an independent router. The main model maintains only a controllable set of core tool definitions, preserving a stable request prefix; tasks that the core tool set cannot cover are handled by invoking \texttt{INTERNAL\_ROUTER}, which uses an independent routing sub-model to discover and execute long-tail tools. Dynamic tool sets stay out of the main model's request prefix, isolating ecosystem changes from its cache path.

\textbf{(2) A placeholder-based intent triage mechanism (Placeholder Tool Triage).} To further stabilize the tool-definition prefix of the main model's high-frequency requests, this paper introduces a fixed-structure placeholder tool, \texttt{TOOL\_WHICH\_WOULD\_BE\_USED}, in the first-round intent triage stage. The definition of this tool never changes during system operation and its invocation is explicitly forbidden, so the model can perceive the existence of external capabilities in the first-round request without triggering any real tool execution. The placeholder has two functions. As a cache anchor, the placeholder tool's definition is independent of the total number of registered tools; no matter how the tool ecosystem expands, the tool-definition prefix of the first-round request remains structurally constant, providing favorable initial conditions for token-level cache hits. As an intent-guiding signal, the placeholder tool's description summarizes the boundary of external capabilities, guiding the model to separate tasks that require environment interaction from pure knowledge questions, and completing task-type triage without introducing any real tool context.

\textbf{(3) A cache-aware tool routing mechanism based on \texttt{INTERNAL\_ROUTER}.} To resolve the structural conflict between tool-scale growth and traditional progressive disclosure strategies, this paper designs a tool routing subsystem independent of the main inference channel. The subsystem is registered as the \texttt{INTERNAL\_ROUTER} pseudo-tool in the main model's fixed tool view; when the main model judges that the current core tool set is insufficient, it can invoke this pseudo-tool to trigger an independent routing sub-model inference channel. Within this channel, the routing sub-model completes the entire process of tool retrieval, candidate filtering, and tool execution, and returns the final execution result to the main model, while the definitions, descriptions, and parameter information of dynamic tools always propagate outside the main model's fixed tool set. As a result, the main model's tool-definition prefix does not change with the dynamics of the long-tail tool space, and the token-level cache prefix remains stable. At the same time, the system's access to the full tool ecosystem is not restricted by the static main-model view: long-tail tools remain discoverable and callable within the independent routing channel. By decoupling ``delivering tool definitions to the main model'' from ``selecting tools from the tool space,'' the design reconciles the dynamism required by progressive disclosure with the prefix stability required by token-level cache hits. The routing channel is also equipped with a multi-stage filtering pipeline (blacklist, whitelist, Top-$K$ semantic pre-selection, and a yellowlist cache-priority strategy) that adds safety boundaries, access control, and cache-preference tuning in specific deployment scenarios, improving the engineering tunability of the routing mechanism.

\textbf{(4) AST-based automatic tool registration with runtime hot reloading.} This paper builds an automatic registration system for a standardized skill repository. At startup, the gateway scans the tool directory, statically parses the \texttt{argparse} parameter definitions in tool scripts via Python AST, automatically extracts parameter names, types, default values, enum ranges, and required attributes, and generates unified MCP JSON schemas. When static parsing cannot cover certain cases, runtime reflection is used as a supplement. Meanwhile, the system detects repository changes by polling file snapshots and rebuilds the tool registry when changes are found, enabling runtime updates without service restarts. Tools can then be added, modified, and removed in parallel with system operation.

\textbf{(5) A skill bootstrapping mechanism that preserves the main cache path.} This paper further enables agents to build and consolidate new skills incrementally through ad-hoc code execution. When the system encounters a task that the existing tools cannot directly cover, it can leverage its existing execution capabilities to generate new skill scripts and metadata and incorporate them into the dynamic skill repository. New skills by default enter an independent tool package and do not automatically change the tool set visible to the main model; subsequent tasks discover these capabilities on demand through the dynamic routing mechanism. The system thus forms a bootstrapping loop from task requirements to capability generation, tool registration, and later reuse, while avoiding direct disruption of existing cache paths by capability expansion.

The paper proposes no new retrieval algorithm. Its core is a reorganization of tool orchestration around \textbf{request structure} and \textbf{tool life-cycle management}: the main model handles stable reasoning and frequent tool calls, the routing channel handles dynamic discovery, the gateway handles registration and hot reload, and the repository handles persistence. Tool growth no longer implies synchronized growth of the main model's cache context.

\subsection{Paper Organization}

The remainder of this paper is organized as follows. Section 2 reviews related work on tool invocation, tool retrieval, prompt caching, and self-evolving agents. Section 3 presents the overall system design: the five-layer architecture, the three-stage decision flow, and the cache-invariance guarantees. Section 4 formalizes the cache-aware progressive disclosure mechanism, covering the problem formulation, the static core tool view, the \texttt{INTERNAL\_ROUTER} routing channel, and the filtering pipeline. Section 5 reports a prototype-scale validation consisting of eight functional experiments and a 30-turn long-chain dialogue. Section 6 concludes and discusses limitations and future work.

\section{Related Work}

\subsection{Agent Frameworks and Tool Orchestration}

As LLMs moved from pure text generation to external tool invocation, tool use became a standard part of agent systems. Early research focused on how models learn to identify when to use tools, generate correct parameters, and exploit tool-returned results for subsequent reasoning. ReAct (Yao et al. 2023) demonstrated an agentic paradigm in which reasoning traces and actions alternate, showing that reasoning and tool use reinforce each other; ART (Paranjape et al. 2023) combined multi-step reasoning with external tool calls through programmatic prompts. Toolformer (Schick et al. 2023) trained language models in a self-supervised manner to call external APIs, enabling the model to decide autonomously when to invoke a tool, which tool to choose, and how to use the results, showing that coupling language models with external tools is feasible. Gorilla (Patil et al. 2023) further targeted large-scale API invocation scenarios, enhancing the model's capability in parameter generation and API selection through API-document retrieval, and treated retrieval-based adaptation as necessary when tool documentation changes. ToolLLM/ToolBench (Qin et al. 2023) started from a much larger set of real-world APIs and built a complete tool-use framework covering data generation, model training, and evaluation, extending the research object from a few exemplar tools to large-scale real-world APIs.

These works focus on model capability and call correctness, i.e., whether the model understands a tool and fills in its parameters correctly. Our concern starts where tool scale grows: how tools are organized, when they are exposed, and whether tool-set changes affect request caching.

On the engineering side, LangChain (LangChain 2026) and AutoGPT (AutoGPT 2026) provide mature abstractions for building tool-augmented agents, HuggingGPT (Shen et al. 2023) orchestrates multiple models into one pipeline through task planning, and LLMCompiler (Kim et al. 2024) schedules parallel function calls to reduce end-to-end latency.

These harnesses make tool invocation convenient, but the abstraction is still a direct model-to-tool call; when the tool set grows, the application layer injects changing tool lists into the request, which perturbs prefix-based API caching.

Prior work optimizes tool-calling ability and task completion. We study how tool definitions enter the request context and make cache stability an explicit design goal of tool orchestration, a layer that can sit on top of existing frameworks.

\subsection{Tool Retrieval and Progressive Disclosure}

When a tool set reaches hundreds or even thousands of entries, sending all definitions to the model inflates the context and enlarges the model's decision space. Tool retrieval and candidate filtering have become standard components of such systems; a recent survey covers the area (Huang et al. 2024). Existing studies typically compress a large tool space into a small set of candidates relevant to the current task through retrieval augmentation, candidate ranking, or staged search.

Gorilla (Patil et al. 2023) introduced API-document retrieval so that the model can obtain the required information from continuously changing API documentation according to the current task, improving the reliability of API calls. ToolLLM/ToolBench (Qin et al. 2023) further built a large-scale API tool-learning and evaluation system, enabling tool retrieval and invocation to be studied in massive real-API environments; AnyTool (Du et al. 2024) localizes tools over large-scale API collections through hierarchical retrieval and self-reflection, and ToolACE (Liu et al. 2024) constructs a large, high-quality tool set to improve models' function-calling ability. This line of research shows that, once tool scale grows, ``first find the relevant tools in a large tool space, then let the model use them'' becomes a more reasonable strategy than exposing all tools to the model directly.

Another idea closely related to tool retrieval is \textbf{progressive disclosure}. Its core idea is to avoid exposing all tools or capability information to the model at once, and instead gradually narrow the candidate range or add necessary information as the task progresses. Such methods alleviate the context burden of large tool sets and help reduce the model's decision complexity when facing many candidate tools.

However, the main optimization target of existing tool-retrieval and progressive-disclosure methods remains \textbf{tool-selection quality}, such as retrieval recall, tool-call correctness, or task completion rate. For a dynamically changing tool subset $S_t$, these methods usually care about whether the selected tools are relevant enough, but rarely discuss another system-level consequence: \textbf{whether the tool set $S_t$ itself changes the cache prefix of model API requests.}

This distinction matters especially for API services with prompt caching. If tool-selection results differ from round to round, the tool-definition segment also changes. Even if only a few tools are provided to the model in each round, such dynamic changes may prevent the request prefix from being reused continuously, compared with a fixed tool set. Simply reducing the number of tools does not necessarily improve system efficiency; under a cache-billing model, the impact of tool-set changes on request-prefix stability must also be considered.

This paper adds a \textbf{cache-stability} requirement to progressive disclosure. The main model's tool list stays static; everything else is pushed into an on-demand routing channel. Disclosure no longer means modifying the main model's tool list; it means reaching additional capabilities through an independent channel when needed.

The design draws a boundary between retrieval and request construction: retrieval faces the whole dynamic tool space, while the main model's fixed definitions never change because of retrieval results. What separates this from traditional tool retrieval is \textbf{whether filtering changes the main model's request structure}.

\subsection{Prompt Caching and System Design}

As LLM API services scale up, prompt caching has become a standard mechanism for reducing the cost of repeated input processing. Major model providers now offer automated input-caching capabilities. For example, DeepSeek API provides a context-caching mechanism (DeepSeek 2026) that caches and reuses repeated prefixes, and OpenAI also offers prompt caching (OpenAI 2026) to reduce the processing cost of repeated input tokens. Because these mechanisms depend heavily on request prefixes, how a system constructs stable input prefixes has gradually become a practical problem in agent engineering.

Existing caching research and system design mainly fall into two directions. One line of work focuses on KV-cache management in model inference infrastructure (Gim et al. 2023; Kwon et al. 2023; Zheng et al. 2024), including cache sharing, cache reuse, cache eviction, and memory management. The other line focuses on API-level prompt caching, such as how to identify reusable input prefixes and how to reduce the computational and economic cost of repeated requests through caching.

These works optimize the serving infrastructure. For agents, the request prefix is not a static system prompt: dialogue history, tool definitions, and tool results all enter subsequent requests, so the agent's own architecture shapes cache conditions.

In tool-augmented agents this matters most: tool definitions are not just capability descriptions, they are part of the request input. Dynamically adding or removing tools is therefore equivalent to rewriting the request structure at runtime, which turns cache hit rate from a serving-side metric into a design constraint.

Our perspective differs from conventional cache optimization, which asks how to make issued requests hit the cache better. We instead design request paths so that high-frequency requests naturally have stable prefixes.

We treat the tool set as part of the request structure: fixing the core set and isolating long-tail tools prevents ecosystem changes from propagating to the main request path, a change made at the application layer rather than inside the caching strategy.

The contribution is a harness architecture that treats cache stability as a design constraint at the tool-orchestration stage; no new caching algorithm is introduced.

\subsection{Self-Evolving Agents}

A second line of recent work lets agents acquire new capabilities at runtime, moving from static tool use toward dynamic skill accumulation and self-extension.

Voyager (Wang et al. 2023) is a representative work in this direction. Targeting the Minecraft environment, it combines an automatic curriculum, an executable skill library, and iterative prompting that leverages environment feedback and execution errors, allowing the agent to keep exploring the environment, generating new code skills, and storing them in the skill library. The skill library can be retrieved and reused in later tasks, forming a continuously accumulating capability system.

Generative Agents (Park et al. 2023) demonstrate from a different angle how long-running agents form sustained behavioral capabilities through memory, reflection, and dynamic retrieval. By recording agent experiences, forming high-level reflections, and dynamically retrieving relevant memories during later planning, the agents accumulate information across sustained interactions and use it to shape future behavior.

Along the same lines, Tool Maker (Cai et al. 2023) further shows that language models can autonomously generate, validate, and reuse tools during operation, moving tool use from ``calling external capabilities'' toward ``manufacturing tools on their own.'' These works show that an agent's capabilities need not rely entirely on the model parameters or tool sets fixed at deployment time; they can be expanded continuously through the code, skills, and memories produced during operation. For this paper, this idea corresponds directly to the bootstrapping evolution of the tool repository: when existing tools cannot satisfy task requirements, the system can construct new skills using its existing execution capabilities and consolidate them into reusable tools.

Self-evolution creates a stability problem of its own: if every generated tool is added to the main model's tool list, its request prefix keeps changing. Coverage improves, but existing cache prefixes break; this is the same conflict as dynamic retrieval.

Generated skills land in an independent package (e.g., \texttt{other}) instead of the main model's fixed set; the gateway registers them automatically, and the router reaches them later. The skill-generation process never modifies the main model's fixed tool view.

Skill bootstrapping, tool registration, and model caching thus form a continuous system loop:

new task creates a capability gap $\rightarrow$ ad-hoc execution or skill generation $\rightarrow$ skill consolidation into the repository $\rightarrow$ automatic gateway registration $\rightarrow$ on-demand discovery by the router $\rightarrow$ reuse in subsequent tasks.

Compared with Voyager's internal code fragments, our generated skills are standard MCP skill packages that enter the tool ecosystem directly and, being outside the main model's visible set, never touch its cache prefix.

\subsection{Summary and Research Positioning}

Viewed through the lens of tool delivery, existing approaches fall into three camps. \textbf{(A) Expose all tools in every request}, the default in early tool-use systems and most agent frameworks: coverage is complete and the request prefix is fixed, but context cost grows linearly with the tool count. \textbf{(B) Retrieve a relevant subset per query} (Gorilla, ToolLLM, AnyTool; progressive disclosure generally): context stays small, but the tool segment changes between requests, undermining prefix-based caching. \textbf{(C) Optimize prompt caching at the serving layer} (DeepSeek, OpenAI, KV-cache work): caching is a service property, and the agent's tool-orchestration behavior is outside the design. Each camp solves one side of the trade-off; none treats request-prefix stability as a tool-orchestration constraint.

CacheRouter takes the stable prefix of (A), the context efficiency of (B), and the serving-layer assumptions of (C): the main model always sees the same fixed core set, long-tail tools are retrieved on demand inside an isolated routing channel, and no new caching mechanism is required from the provider. The difference is structural rather than parametric: we decouple tool selection from tool delivery instead of tuning either side in isolation, and the design sits on top of existing frameworks, models, and providers.

\section{System Architecture Overview}

\subsection{Design Principles}

The architecture must satisfy three requirements at once: new tools placed in a directory register themselves automatically and become usable immediately; the request prefix stays stable as the tool count $M$ grows; and the deployed agent can write, validate, and consolidate new capabilities without breaking existing cache patterns. The first three layers (application, decision, routing) organize the reasoning path; the last two (gateway, repository) handle tool delivery and evolution.

The system design follows four basic principles: information locality, forward compatibility, separation of concerns, and runtime hot reloading.

\textbf{Information locality.} Tools are layered by usage frequency; high-frequency core tools and rarely used long-tail tools live in separate request contexts, keeping high-frequency requests small and stable.

\textbf{Forward compatibility.} Adding tools must not change existing request contexts or destroy cached prefixes, so historical cache results keep paying off.

\textbf{Separation of concerns.} Tool selection is decoupled from tool execution, so selection-side changes cannot perturb the execution path or its cache.

\textbf{Runtime hot reloading.} Tools can be added, removed, and updated without restarting the core service: the gateway re-scans the repository and atomically replaces the tool registry. The current implementation and experiments validate the reloading mechanism itself; stronger guarantees about the consistency of in-flight requests under concurrent updates are left for future work.

Together these principles give a dynamic tool ecosystem a stable request architecture: extensible in function, yet stable in context and cache behavior on high-frequency paths.

\subsection{Overall Framework}

\subsubsection{Overview of the Five-Layer Framework}

The system is organized into five layers: application (user interaction), decision (task reasoning), routing (tool filtering), gateway (tool services), and repository (tool resources). Each layer encapsulates complexity downward and offers stable interfaces upward, so tool-set changes stay in the lower layers and cannot disturb the core request path.

\begin{figure}[htbp]
    \centering
    \setlength{\abovecaptionskip}{2pt}
    \setlength{\belowcaptionskip}{2pt}
    \includegraphics[width=0.8\textwidth, trim={137bp 81bp 135bp 43bp}, clip]{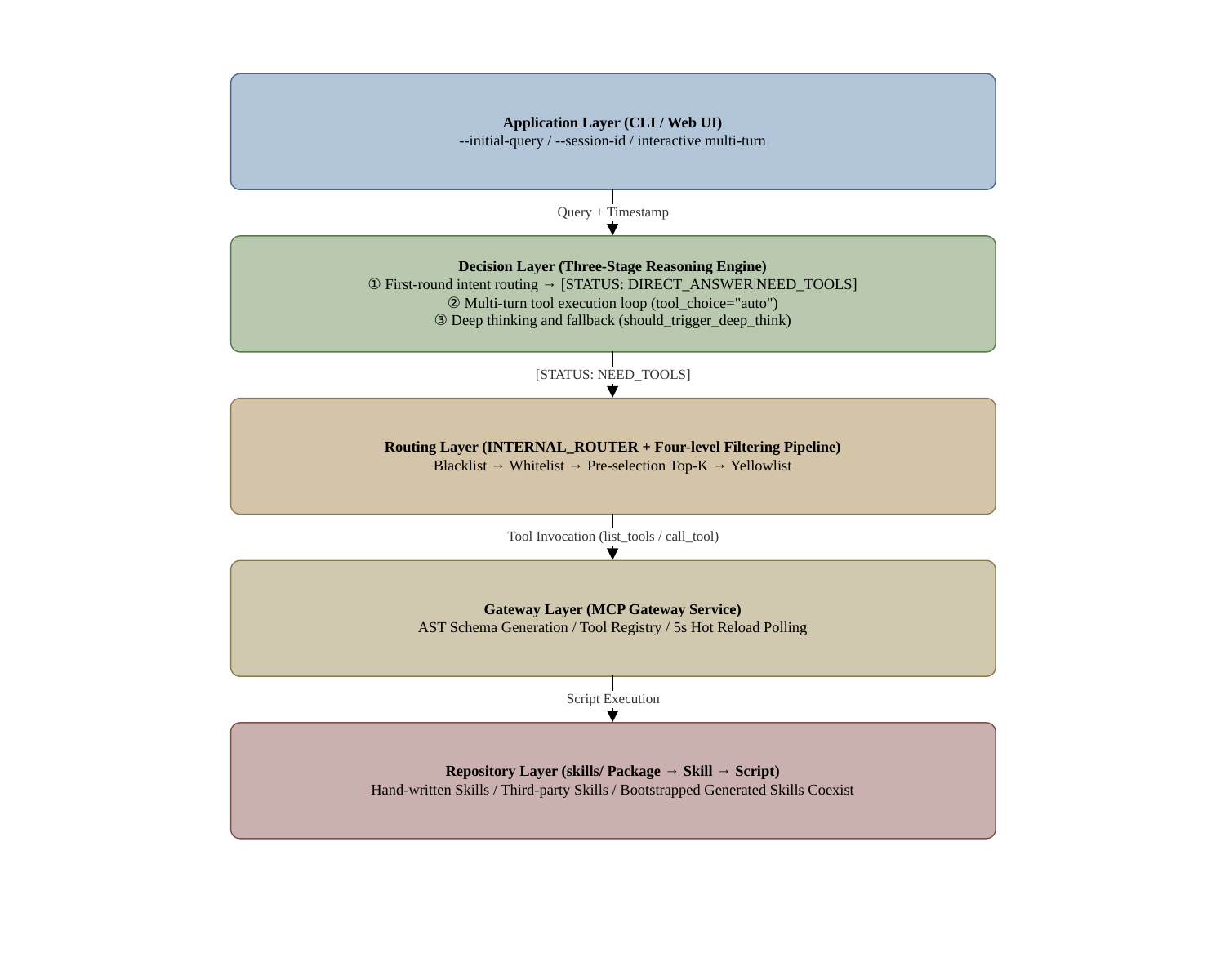}
    \caption{Five-layer architecture. Each layer encapsulates complexity downward and provides service abstractions upward; inter-layer communication is entirely based on the MCP protocol or inter-process parameter passing, with no shared internal state.}
    \label{fig:architecture}
\end{figure}

\textbf{Application layer.} Handles user interaction and session state through a CLI client and a Web UI, and forwards requests to the decision layer without touching tool discovery or execution.

\textbf{Decision layer.} Decides whether a task needs tools and runs the three-stage flow: first-round triage, the tool execution loop, and deep thinking with fallback. Separating the stages keeps the full tool set out of most requests and keeps high-frequency paths stable.

\textbf{Routing layer.} Exposes \texttt{INTERNAL\_ROUTER} as a pseudo-tool: routing logic plus a sub-model narrow the candidates (blacklist, whitelist, Top-$K$, yellowlist), and the main model receives only the structured result, never the long-tail definitions.

\textbf{Gateway layer.} Maintains the tool registry on top of the MCP Gateway Service: parses tool scripts with AST to generate schemas, and polls the tool directory so new, removed, or updated tools take effect without restarts. Upper layers never see the internals of tool scripts.

\textbf{Repository layer.} Stores tools as packages, skills, and scripts with metadata; may hold hand-written, third-party, or runtime-generated skills. It only supplies structured resources, so the tool set can evolve without touching upper layers.

The layers communicate through MCP or inter-process parameter passing and share no internal state, so tool-set changes stay isolated. This is the foundation for stable caching, efficient selection, and runtime tool expansion.

\subsubsection{Application Layer and Decision Layer}

The application layer handles interaction and session state for the CLI and the Web UI. Between the application and decision layers, two history streams are maintained: a compressed stream (system prompt plus one summary per question--answer round) used only by first-round triage, and a full stream (all messages, tool calls, and tool results) used by the execution loop. This keeps the high-frequency first-round path small while preserving complete traceability for complex tasks.

The decision layer runs three stages, summarized in Algorithm 1: first-round triage, the tool execution loop, and deep thinking with fallback. In triage, the model sees only the fixed placeholder tool \texttt{TOOL\_WHICH\_WOULD\_BE\_USED} (its invocation is blocked via \texttt{tool\_choice: "none"}) and appends a \texttt{NEED\_TOOLS} or \texttt{DIRECT\_ANSWER} marker to its answer. The placeholder serves two purposes: as a cache anchor, its definition is independent of the tool count $M$, so the first-round tool segment never changes; as an intent-guiding signal, its description sketches the boundary of external capabilities (real-time data, code execution, file access), so the model can triage without seeing any real tool. The two runtime artifacts are reproduced below.

In the execution loop, the main model sees only the core set plus \texttt{INTERNAL\_ROUTER} and may issue parallel tool calls; results are appended to the full history. Explicit errors (structured error bodies) and implicit errors (keyword patterns for permission denials, missing resources, unavailable tools) do not stop the loop: the model receives a one-round penalty and replans, and unmatched tool calls in the message sequence are completed with placeholder error responses to keep the API message ordering valid.

Deep thinking is triggered by any of three signals (status anomaly, e.g., no direct-answer state or the round limit being reached; time anomaly, i.e., execution exceeding 60 s; or call fragmentation, i.e., four or more consecutive tool-call rounds) and injects a re-planning instruction into the dialogue so the model re-examines the execution trajectory and re-enters the loop. If re-execution still fails, fallback generation answers from the information already available and explicitly marks uncertainty, instead of fabricating content.

Together, the two layers form the core request-processing chain: triage separates simple from tool-dependent tasks, the loop and error handling keep execution convergent, and the staged design never exposes the full tool set or full history in one request, which underpins the routing layer and cache stability on high-frequency paths.

To make the first-round mechanism concrete, we reproduce the two runtime artifacts it relies on: the instruction appended to the system prompt, and the placeholder tool exposed to the model.

\begin{lstlisting}
[First-round instruction, appended to the system prompt]
You have access to a general-purpose external capability tool.
Do not explain the tool usage; just respond normally and append the marker.
When you need to rely on external systems, you MUST break down the user's
problem into at least two sub-tasks, provide a clear step-by-step reasoning
outline, and explicitly analyze which parts require external calls.
[Hard Rule] If the user explicitly says 'call tools' or 'use tools', you
must append [STATUS: NEED_TOOLS] at the end of your response.
\end{lstlisting}

\begin{lstlisting}
TOOL_WHICH_WOULD_BE_USED (placeholder tool, first round only)
description: "Placeholder for general external capability. Represents any
  operation performable via external tools or code execution: fetching
  real-time data (weather, news, ...), executing Python code, reading/writing
  files, calling external APIs, automating tasks. In the first round you only
  need to judge whether external systems are needed; the specific tool calls
  are handled automatically by the subsequent pipeline."
parameters: { "intent_summary": { "type": "string", "required": true } }
\end{lstlisting}

\subsubsection{Routing Layer}

The routing layer keeps the main model's context stable while finding the tools a task needs. It decouples tool selection from tool delivery by registering \texttt{INTERNAL\_ROUTER} as a pseudo-tool in the main model's view: invoking it enters an independent sub-model channel that retrieves, filters, and calls long-tail tools, and the main model receives only the structured result. The tool space becomes a resource pool searched inside the router, so tool growth no longer implies linear growth of the main model's context.

A routing call works as follows: the current dialogue is compressed to a short task summary; the full registered tool list is fetched from the gateway (the router is not bound by the main model's package visibility); a four-stage filter (blacklist, whitelist, Top-$K$ relevance, yellowlist) narrows the candidates; and the routing sub-model makes one constrained tool call. The result is serialized back into the main model's context as plain information; the main model never sees the candidates or the filtering rationale.

Operationally the router is stateless (input is the compressed context plus the current tool list), lightweight (one local tool call, no long-term planning), and deliberately lossy: context compression, candidate truncation, and priority filtering discard tool information in exchange for lower overhead and a more stable main request path.

By combining an independent routing sub-model with a multi-stage filtering mechanism, the routing layer controls how long-tail tool information enters the main model while preserving the ability to discover the full tool set. Tool-scale expansion thus does not directly translate into expansion of the main model's request context; this layer is where cache stability, tool extensibility, and tool-selection efficiency are balanced.

\subsubsection{Gateway Layer and Repository Layer}

The gateway layer wraps heterogeneous skill scripts into unified MCP tool interfaces and provides automatic registration, metadata indexing, and runtime hot reloading to the upper layers.

At startup the gateway scans the repository and registers compliant scripts. Schemas are generated by AST-based static parsing of the script source (no tool code runs during registration), with runtime reflection as a fallback. Global naming and description-length limits keep tool metadata compact enough for router retrieval.

Package-level tool indices let the router pre-select without parsing scripts in real time; index updates are atomic, and orphan indices are cleaned up when packages are removed. Periodic file snapshots trigger re-scanning and schema construction under lock protection, then atomically replace the registry, so tools change at runtime without restarts.

The repository organizes skills in a three-level hierarchy (package, skill, script), separating implementation, description, and classification. It holds hand-written, third-party, and runtime-generated skills; generated skills default to an independent package so they never require changes to the client or gateway.

Package granularity also defines the main model's visibility boundary: stable tools are aggregated into visible packages, while new or low-frequency tools land elsewhere, so their changes never touch the cache prefix. Repository and router together implement ``resource pool, visible boundary, on-demand disclosure.''

Together they form the system's dynamic tool infrastructure, separating how tools are stored and registered from when and which tools the upper layers decide to use.

\subsection{Core Data Flow}

A user query goes through intent triage, tool discovery and execution, error recovery, and state persistence. Depending on triage, it takes either the direct path (answer immediately) or the tool path (gateway startup, tool execution, recovery, persistence); the two paths split at the first-round triage.

The application layer normalizes the query into the compressed history, and the main model performs triage with only the placeholder tool; no real tool is called at this stage. A direct answer is returned immediately with a single API call and a stable tool prefix.

For the tool path, the MCP gateway starts and registers tools; the main model then sees only the core packages (\texttt{--packages-agent}) plus \texttt{INTERNAL\_ROUTER}. Pre-selection and filtering happen inside the routing subsystem and never change what the main model sees.

In the execution loop, ordinary tool calls run concurrently, \texttt{INTERNAL\_ROUTER} calls go to the routing sub-model channel, and after each round the system checks for a direct-answer state; the loop ends on completion or when the round limit is reached.

If the loop does not converge, status/timeout/fragmentation signals trigger deep thinking, which re-plans from the existing trajectory and re-enters the loop; persistent failure leads to a fallback answer that explicitly marks uncertainty. Completed sessions are persisted for recovery, and the gateway keeps hot-reloading tools.

The flow forms a closed loop in which triage selects the path, the main model plans, the router discovers, the gateway executes, deep thinking recovers, and persistence preserves state, separating high-frequency direct requests from complex tool requests early, and keeping tool visibility and context under control.

\subsection{Stability Guarantee: Cache Invariance}

Cache hits here refer to token-level prefix reuse: an identical input prefix is processed once and reused, which is different from the same query returning the same result. When tool definitions or other leading content change, existing caches stop matching even if the task is unchanged, and naively injecting the whole tool set makes the request prefix move with the tool ecosystem. Our fix is to keep the main model's resident tool set fixed, isolate long-tail tools, and keep selection independent, so the ecosystem can evolve while the core request path stays stable.

The payload of the $i$-th main-model request is the system prompt, the message history, and the tool definitions of the core set $C$. If $C$ and the system prompt stay constant, cache hits reduce to a consistency requirement on the history prefix, which append-only growth satisfies naturally: later requests keep the existing prefix and add new messages at the end.

Three runtime events could in principle change $C$; the design keeps it fixed in each case.

\textbf{Adding tools at runtime.} The gateway hot-reloads repository changes, but the main model only sees the preset resident packages; a new tool outside those packages never enters its requests.

\textbf{Generating new skills.} Bootstrapped skills are written into an independent package outside the main model's visible boundary and are discovered by \texttt{INTERNAL\_ROUTER} only when a later task needs them; bootstrapping expands the potential tool space, not the resident context.

\textbf{Restarting the gateway.} The visible tool set is determined by the configured packages, not by the gateway's transient state, so a restart or failure recovery leaves the main model's tool definitions unchanged.

Together, these cases show that the registry and the router may change freely while $C$ stays closed and stable: tool changes first occur in the space invisible to the main model and reach it only as routing results, never as tool definitions.

This is a boundary between a dynamic tool space (extensible, updatable) and a stable reasoning space (fixed core tools and system prompt), connected by \texttt{INTERNAL\_ROUTER}: long-tail tools are accessible when needed, but tool-count growth never translates linearly into the main model's request size.

Cache invariance does not require identical payloads; it only requires that an established prefix is not invalidated by irrelevant changes. History keeps growing, but dynamic tool changes do not break it: the goal is to avoid prefix invalidation caused by tool-ecosystem changes, not to stop context growth.

A fixed core view also pays for itself: with \(s_{\mathrm{avg}}\) tokens per tool definition, full loading costs \(O(M \cdot s_{\mathrm{avg}})\) while the static core set costs \(O(|C| \cdot s_{\mathrm{avg}})\) with \(|C| \ll M\), and because the tool segment is identical across requests, the one-time cache-establishment cost is amortized over many calls. The trade-off is that tools in \(T \setminus C\) (code execution, complex file operations, domain-specific queries) are not directly callable; the routing channel of Section 4 restores that coverage on demand, so expansion happens outside the main model's cache path. Costs also shift by usage frequency: high-frequency requests carry the stable cost of \(C\), while long-tail tools are paid for only when actually needed instead of being amortized over every request.

\section{Cache-Aware Progressive Disclosure}

To resolve the conflict between tool-scale expansion and model-request cache stability, this paper proposes a cache-aware progressive disclosure mechanism. Its core idea is to relieve the main model of the full tool set in every request: instead of reconstructing the complete visible set for each query, the tool space is partitioned into a stable core tool space and a dynamic long-tail tool space. High-frequency, fundamental, and relatively stable tools form the main model's fixed tool set and are invoked directly by the main reasoning channel; low-frequency, specialized, or dynamically generated long-tail tools remain in an independent tool space and are retrieved and invoked on demand through the routing subsystem only when a task actually requires them.

The mechanism decouples ``tool selection'' from ``tool-definition delivery'': the main model keeps a stable tool view of controllable size instead of searching the entire tool space directly, while dynamic discovery and execution happen in an independent routing channel. Access to the full tool ecosystem is retained without expanding the main model's request context, balancing extensibility, context efficiency, and cache stability.

\subsection{Problem Formulation}

The core problem can be summarized as a structural conflict between tool-selection flexibility and API cache efficiency. Throughout the paper, ``cache hit'' refers to token-level prefix reuse: an identical input prefix is processed once and reused, not the same query returning the same answer. The dynamism required by progressive disclosure, as discussed in Section 1.1, constitutes the core contradiction that CacheRouter must reconcile. As the number of tools grows, injecting all tool definitions into the main model provides high tool coverage but incurs large context overhead; dynamically selecting tools according to the user query reduces irrelevant tool information but may cause tool definitions to differ across requests, weakening request-prefix stability.

Let \(T\) denote the complete set of registered tools, with total size \(M\). For the \(i\)-th request, the tool set directly visible to the main model is denoted as \(S_{\mathrm{agent}}^{(i)} \subseteq T\). The effective prefix of the main model request can be expressed as
\begin{equation}
P_i = (\mathrm{system\_prompt} \parallel \mathrm{history}_i \parallel \mathrm{tools}(S_{\mathrm{agent}}^{(i)})),
\end{equation}
where \(\parallel\) denotes concatenation in a fixed order and \(\mathrm{history}_i\) is the historical context of the \(i\)-th request. For caching mechanisms that rely on prefix reuse, the tool-definition set determines the model's tool-selection space and, just as directly, the stability of the request cache structure.

To resolve the contradiction, this paper adopts a decoupled design. Let the core tool set fixed and visible to the main model be \(C \subset T\), with \(|C| \ll M\). The main model always maintains this stable tool view \(C\), while long-tail tools in \(T \setminus C\) are accessed on demand through an independent routing channel. The core request of the main model can then be expressed as
\begin{equation}
P_C = (\mathrm{system\_prompt}_C \parallel \mathrm{history}_i \parallel \mathrm{tools}(C)).
\end{equation}
Since the tool-definition segment is determined by the fixed set \(C\), dynamic changes of long-tail tools do not directly enter the main model's fixed request prefix. At the same time, the system can still access tools in \(T \setminus C\) through the independent router, so overall tool coverage is not sacrificed by the static main-model view.

The system forms two complementary processing channels: the main channel handles high-frequency core tools and main-task reasoning, favoring stability and cache reuse; the routing channel handles long-tail discovery and execution, favoring dynamism and coverage. Algorithm~\ref{alg:main} summarizes the resulting pipeline.

\begin{algorithm}[htbp]
\caption{Query handling pipeline}
\label{alg:main}
\begin{algorithmic}[1]
\Require query $q$; compressed history $H_c$; full history $H_f$; core tool set $C$; router pseudo-tool $R$; round limits $R_1 < R_2$
\State first round: $\langle H_c,\ q,\ \mathrm{tools}=\{\mathrm{TOOL\_WHICH\_WOULD\_BE\_USED}\},\ \mathrm{tool\_choice}=\mathrm{none}\rangle$
\If{the answer carries the \textsc{direct\_answer} marker}
  \State \Return the answer
\Else
  \State $r \gets 0$; $\mathit{limit} \gets R_1$
  \Repeat
    \State send $\langle H_f,\ q,\ \mathrm{tools}=C \cup \{R\}\rangle$
    \If{the model calls a core tool $t \in C$}
      \State execute $t$; append the result to $H_f$
    \ElsIf{the model calls $R$}
      \State append $\Call{Router}{q, H_f}$ to $H_f$ \Comment{Algorithm \ref{alg:router}}
    \ElsIf{the model returns a final answer}
      \State \Return the answer
    \EndIf
    \State $r \gets r + 1$
    \If{a deep-thinking signal fires (status anomaly, $>60$~s, or $\geq 4$ tool-call rounds)}
      \State run deep thinking; $\mathit{limit} \gets R_2$
    \EndIf
  \Until{$r \geq \mathit{limit}$}
  \State \Return the fallback answer
\EndIf
\end{algorithmic}
\end{algorithm}

\subsection{Dynamic Tool Routing: \texttt{INTERNAL\_ROUTER}}

To recover the coverage lost to the static view, we register a meta-tool, \texttt{INTERNAL\_ROUTER}, in the main model's fixed tool view. It carries no business logic; it is the entry point to the dynamic tool space.

When $C$ is insufficient, the main model invokes it; an independent sub-model channel retrieves, filters, and executes the right long-tail tool and returns the result.

From the main model's side, the tool set stays $C \cup \{\texttt{INTERNAL\_ROUTER}\}$ regardless of $M$; dynamic tool information propagates only inside the routing subsystem.

This is progressive disclosure in its purest form: a stable core view first, with additional tool information disclosed to the router only when the task demands it.

\subsubsection{Routing Protocol}

The execution process of \texttt{INTERNAL\_ROUTER} consists of five stages: context compression, tool-list refresh, candidate filtering, routing sub-model reasoning, and result return.

The five stages are condensed in Algorithm 2. The dialogue is compressed into a task summary of at most 800 characters; the full tool list is fetched from the gateway (the router is not bound by the main model's package visibility, so it can see all registered tools, including newly added and generated ones); the four-stage filter narrows it to at most \(k=5\) candidates; the routing sub-model makes one tool call following the ``fast tool caller'' prompt, with a 2000-token generation limit; and the result is serialized into a JSON string and returned to the main model. Dynamic tool information never leaves the routing subsystem in the form of tool definitions.

The router is stateless, lightweight, and deliberately lossy, as discussed in Section 3.2.3. Algorithm~\ref{alg:router} condenses the protocol into operational steps, and the system prompt of the routing sub-model is reproduced below.

\begin{algorithm}[htbp]
\caption{\textsc{Router} sub-channel (triggered by \texttt{INTERNAL\_ROUTER})}
\label{alg:router}
\begin{algorithmic}[1]
\Require task summary $s$ ($\leq 800$ chars); full gateway tool list $T$; blacklist $B$; whitelist $W$; yellowlist $Y$; candidate size $k = 5$
\State $L \gets \Call{list\_tools}{}$ \Comment{entire registry, not limited to $C$}
\State $L \gets B(L)$ \Comment{hard block}
\If{$W \neq \emptyset$}
  \State $L \gets W(L)$ \Comment{restrict to whitelist}
\EndIf
\State $L \gets \Call{TopK}{L, q, k}$ \Comment{keyword relevance pre-selection}
\State $L \gets Y(L)$; if $L = \emptyset$, keep the top-1 candidate
\State sub-model request: $\langle s,\ q,\ \mathrm{tools}=L,\ \mathrm{tool\_choice}=\mathrm{auto}\rangle$
\If{the sub-model selects a tool $t$}
  \State execute $t$; \Return the JSON result
\Else
  \State \Return a retry signal
\EndIf
\end{algorithmic}
\end{algorithm}

The routing sub-model is instructed with the following ``fast tool caller'' prompt:

\begin{lstlisting}
You are a fast tool caller. Your primary responsibility is to select the most
appropriate tool from the given list and call it to fulfill the user's request.
[Important Rules]
1. You should call a tool whenever possible. Call only ONE tool per request.
2. If none of the available tools can reasonably fulfill the request, clearly
   explain why and suggest alternatives, rather than forcing a call.
3. After calling a tool, return the result directly. Do not explain your
   reasoning or the process.
4. If the tool returns an error, return the error message as-is.
5. Do not fabricate data or make assumptions beyond what the tools provide.
\end{lstlisting}

\subsubsection{Multi-Stage Filtering Pipeline}

To keep the routing channel from facing all $M$ tools, filtering runs in four stages that layer safety, deployment, relevance, and cache concerns.

Let the initial tool universe be \(T^{(0)} = T\). After the four stages, the tool set shrinks progressively and finally satisfies

\begin{equation}
|T^{(4)}| \leq k = 5.
\end{equation}

The first stage is blacklist filtering. Tools specified by \texttt{ROUTER\_BLACKLIST} as forbidden for the router are unconditionally removed. This stage is a hard constraint, mainly for safety risks or capabilities explicitly forbidden for automatic invocation in deployment environments.

The second stage is whitelist filtering. When \texttt{ROUTER\_WHITELIST} is non-empty, only the allowed tools are retained, satisfying tool-access restrictions in specific deployment scenarios. When the whitelist is empty, no additional restriction is imposed.

The third stage is Top-\(K\) pre-selection. The system ranks the remaining tools via \texttt{select\_tools(query, tools, INITIAL\_TOP\_K)}, combining tool names, descriptions, parameter specifications, module documentation, function documentation, and \texttt{SKILL.md} content, using the hit ratio of query terms in tool metadata as the relevance score. This stage mainly performs candidate compression rather than final decision-making, so a small loss of search precision at low computational cost is acceptable.

The fourth stage is yellowlist filtering. Yellowlist filtering brings cache benefits into tool selection. The system automatically adds the tools in the main model's resident set \(C\) to a soft-avoidance set, preferring to retain long-tail tools outside the core tool set in dynamic routing candidates. When the main model can already complete the task by invoking tools in \(C\) directly, that path is preferred and unnecessary routing sub-model calls are avoided.

However, the yellowlist is not an absolute prohibition. If all remaining candidates belong to \(C\), the system still retains the most relevant tool as a fallback candidate to avoid tool unavailability caused by cache optimization. This stage embodies ``cache-priority'' rather than ``cache-absolute-priority.''

The four stages form a progressive chain (safety, deployment, relevance, cache preference), each bearing a distinct concern, so routing behavior stays interpretable instead of being compressed into a single model judgment.

The four-stage filtering can also be written as a layer-by-layer shrinking process. Let \(F_B\), \(F_W\), \(F_K\), and \(F_Y\) denote the blacklist, whitelist, Top-\(K\) pre-selection, and yellowlist filtering operators, respectively. The final candidate set can be written as
\begin{equation}
T^{(4)} = F_Y(F_K(F_W(F_B(T)))),
\end{equation}
where each filtering function operates on the candidate set output by the previous stage. The expression reflects the composition of different constraints: later stages do not re-expand the tool space already excluded by earlier stages.

\subsubsection{Dual-Layer Cache Compatibility Analysis}

The static core set and the dynamic routing channel create two request domains with different stability goals: the main-model domain prioritizes prefix stability and cache reuse, the routing domain prioritizes access to a changing tool space. Their compatibility follows from three properties.

\textbf{Prefix stability.} Router-side refresh, filtering, and execution never write back into the main model's tool definitions: main-model requests consist of the system prompt, the history, and \(\mathrm{tools}(C)\), so \(\mathrm{tools}(C)\) is unchanged however the dynamic space evolves. With append-only history growth, later requests keep the existing prefix and append new messages, so the structural condition for reuse persists. This is an architectural guarantee about the request structure, not a promise about the actual hit rate, which also depends on server-side eviction and cache life cycles.

\textbf{Bounded routing load.} With \(k=5\) candidates and \(r_{\max}=3\) routing invocations per round, the tool-definition overhead of one round is \(O(15 \cdot s_{\mathrm{avg}})\) tokens, independent of the total tool count \(M\) (versus \(O(M \cdot s_{\mathrm{avg}})\) for full exposure). Dynamic access is thus bounded and paid for only by requests that actually need long-tail tools.

\textbf{Domain isolation.} Main-model and routing requests have different system prompts, context sources, and tool sets, so dynamic changes affect routing requests without propagating to main-model requests. The two domains still interact: the router's result returns as a JSON string, but what crosses the boundary is task-result information, not tool definitions.

The main model carries the high-frequency, stable cache path, while dynamic routing carries the low-frequency, changing tool-access path. The system does not require identical cache states across requests; structural isolation simply directs cache resources to the path with the highest invocation frequency and the most stable prefix.

\section{Experimental Evaluation}

\subsection{Experimental Setup and Evaluation Metrics}

This section reports a prototype-scale validation of the architecture. Eight functional experiments exercise the main execution paths (multi-step sequential tasks, parallel tasks, long-chain dialogue, task triage, convergence on complex tasks, fallback for unanswerable queries, \texttt{INTERNAL\_ROUTER} triggering and filtering, and skill-bootstrapping consolidation), followed by a dedicated 30-turn dialogue experiment that tracks token-level cache hit rate and multi-turn stability in a single session.

The experiments are based on the CacheRouter system prototype, running in a Linux environment (implemented in Python, with the MCP Gateway Service providing tool registration and execution). Both the main model and the routing sub-model call the DeepSeek API (DeepSeek 2026). Experiments 1--8 contain 55 query cases in total, and the long-chain multi-turn dialogue experiment contains 30 query cases. All experiments are executed sequentially in a single session to cover system behavior under continuously accumulating session history. During Experiments 1--8, the gateway loaded 11 tools distributed across five tool packages: base (3), computer-program (4), energy-preserving-method (2), geography (1), and internet (1); the core tool set fixed and visible to the main model is specified by \texttt{--packages-agent}. The evaluation is a feasibility check of the design, not a benchmark against other systems: the query set is modest, the tasks were chosen to cover each functional path, and the reported numbers are descriptive rather than significance tests.

The evaluation metrics used in this section are defined as follows.

\textbf{Task completion rate:} the proportion of queries that successfully converge to a final answer.

\textbf{Execution rounds:} the number of model reasoning rounds produced by a single query in the tool execution loop; direct-answer tasks are counted as 0 rounds.

\textbf{Number of Router invocations:} the total number of times the \texttt{INTERNAL\_ROUTER} pseudo-tool is invoked by the main model, reflecting the usage frequency of the long-tail tool routing channel.

\textbf{Per-query latency:} the wall-clock time from a query entering the system to the generation of the final answer.

\textbf{Token-level cache hit rate:} the core evaluation metric of this paper. All cache-hit data are computed at token granularity rather than query granularity. It is defined as
\begin{equation}
\text{Token-level cache hit rate} = \frac{\text{cached input tokens}}{\text{cached input tokens} + \text{uncached input tokens}}.
\end{equation}
The data are taken from the official usage statistics fields \texttt{prompt\_cache\_hit\_tokens} and \texttt{prompt\_cache\_miss\_tokens} of the DeepSeek API and cover all API calls, including every round in the tool execution loop and routing sub-model requests. For cost accounting, we follow DeepSeek's official pricing documentation, where the billing ratio between cache-hit and cache-miss input tokens is about 1:30. Given a hit rate $h$, the input cost after caching is approximately $((1-h)\times 30 + h)/30 = (1-h) + h/30$ of the no-cache baseline.

\subsection{Experiments 1--8: Functional and Stability Evaluation}

Experiments 1--8 exercise the functional paths described above; their token-level cache statistics are aggregated at the end of this subsection.

\subsubsection{Experiment 1: Multi-Step Sequential Tasks}

Experiment 1 checks completion rate and execution-chain stability for tasks that require strictly sequential tool calls. Five queries form serial chains such as ``query weather and save to a file'' and ``RAG retrieval and write to a file''; Table \ref{tab:exp1} lists the per-case results.

\begin{table}[htbp]
    \centering
    \caption{Experiment 1: Results of multi-step sequential tasks}
    \label{tab:exp1}
    \begin{tabular}{lcccp{6.2cm}}
        \hline
        ID & Rounds & Router & Time (s) & Answer summary \\
        \hline
        1.1 & 11 & 5 & 150.8 & Beijing weather query, saved to file and verified via Router (switched to /tmp sandbox path after path restriction) \\
        1.2 & 3 & 0 & 12.3 & rag\_project1 retrieved campus, 1 hit \\
        1.3 & 4 & 0 & 23.3 & Beijing 31.5°C vs. Shanghai 29.4°C, Beijing hotter \\
        1.4 & 10 & 6 & 115.7 & faculty retrieved 0 hits, wrote to file via Router (succeeded after whitelist restriction) \\
        1.5 & 4 & 1 & 26.6 & Beijing weather vs. campus retrieval, structured output \\
        \hline
    \end{tabular}
\end{table}

All five queries completed successfully. When the deployment whitelist blocked file paths (1.1, 1.4), the model found the /tmp sandbox via \texttt{INTERNAL\_ROUTER}, replanned, and converged; the routing channel adapts to execution-environment constraints.

\subsubsection{Experiment 2: Parallel Tasks}

Experiment 2 tests concurrent execution of independent tool calls. Five queries cover parallel weather queries for several cities and parallel retrieval across multiple RAG projects; results appear in Table \ref{tab:exp2}.

\begin{table}[htbp]
    \centering
    \caption{Experiment 2: Results of parallel tasks}
    \label{tab:exp2}
    \begin{tabular}{lcccp{6.2cm}}
        \hline
        ID & Rounds & Router & Time (s) & Answer summary \\
        \hline
        2.1 & 11 & 0 & 25.7 & Parallel Beijing/Shanghai weather queries, including humidity and dew point details \\
        2.2 & 3 & 0 & 19.1 & Parallel retrieval across two RAG projects (BUPT + Mathematics), both successful \\
        2.3 & 4 & 0 & 17.7 & Beijing weather and rag\_project1 in parallel \\
        2.4 & 4 & 0 & 25.2 & Shanghai weather and rag\_project2 in parallel (doctoral retrieval succeeded) \\
        2.5 & 13 & 0 & 68.3 & Parallel weather queries for three cities, summarized \\
        \hline
    \end{tabular}
\end{table}

All five queries succeeded; the parallel calls ran concurrently (case 2.5 finished a three-city summary in 68.3 s), and 0 Router invocations confirm that the core set covers such high-frequency tasks.

\subsubsection{Experiment 3: Long-Chain Dialogue and Cache Stability}

Experiment 3 observes cache-hit behavior in a continuous dialogue to check request-prefix stability. The 10 queries are divided into groups A and B: group A keeps the gateway resident, and group B cold-starts the gateway process before each group of queries, comparing the impact of gateway life-cycle changes on request prefixes. Table \ref{tab:exp3} lists the per-case results.

\begin{table}[htbp]
    \centering
    \caption{Experiment 3: Results of long-chain dialogue}
    \label{tab:exp3}
    \begin{tabular}{lcccc p{5.2cm}}
        \hline
        Group & ID & Rounds & Router & Time (s) & Answer summary \\
        \hline
        A & Q1 & 4 & 0 & 15.7 & Beijing weather \\
        A & Q2 & 3 & 0 & 18.2 & rag\_project1 BUPT retrieval \\
        A & Q3 & 3 & 0 & 11.1 & Shanghai weather \\
        A & Q4 & 3 & 0 & 10.1 & rag\_project2 Mathematics retrieval \\
        A & Q5 & 3 & 0 & 9.3 & Guangzhou weather \\
        B & Q1 & 3 & 0 & 14.3 & Beijing weather (cold start) \\
        B & Q2 & 3 & 1 & 26.6 & rag\_project1 BUPT (cold start) \\
        B & Q3 & 4 & 0 & 14.1 & Shanghai weather (cold start) \\
        B & Q4 & 3 & 0 & 9.3 & rag\_project2 Mathematics (cold start) \\
        B & Q5 & 4 & 0 & 14.4 & Guangzhou weather (cold start) \\
        \hline
    \end{tabular}
\end{table}

All ten queries completed. In group B, restarting the gateway between query groups left the main model's tool definitions unchanged (Section 3.4), so the long-chain dialogue kept reusing its cache prefix.

\subsubsection{Experiment 4: Triage and Direct Answering of Simple Queries}

Experiment 4 checks whether first-round placeholder triage separates knowledge questions from tool-type queries and keeps simple tasks out of the execution flow. Five queries were used, two pure knowledge questions and three tool-type queries; Table \ref{tab:exp4} lists the per-case results.

\begin{table}[htbp]
    \centering
    \caption{Experiment 4: Results of simple queries}
    \label{tab:exp4}
    \begin{tabular}{lcccp{6.2cm}}
        \hline
        ID & Rounds & Router & Time (s) & Answer summary \\
        \hline
        4.1 & 0 & 0 & 1.7 & 2 + 2 = 4, direct answer \\
        4.2 & 0 & 0 & 1.6 & The capital of China is Beijing, direct answer \\
        4.3 & 4 & 0 & 15.2 & Beijing weather (tool query) \\
        4.4 & 3 & 0 & 26.5 & rag\_project1 BUPT retrieval succeeded \\
        4.5 & 4 & 0 & 18.4 & Shanghai weather \\
        \hline
    \end{tabular}
\end{table}

All five queries succeeded. The two knowledge questions were triaged as direct answers (0 rounds, about 1.6 s each) and the three tool queries entered the execution loop, matching the placeholder-triage behavior described in Section 3.2.2.

\subsubsection{Experiment 5: Replanning and Convergence on Complex Tasks}

Experiment 5 looks at convergence under frequent tool calls and high task complexity, and at replanning when execution is blocked. Five queries cover a four-city weather comparison and long code generation; Table \ref{tab:exp5} gives the per-case results.

\begin{table}[htbp]
    \centering
    \caption{Experiment 5: Results of complex queries}
    \label{tab:exp5}
    \begin{tabular}{lcccp{6.2cm}}
        \hline
        ID & Rounds & Router & Time (s) & Answer summary \\
        \hline
        5.1 & 15 & 1 & 43.6 & Four-city weather comparison, Beijing 31.5°C hottest \\
        5.2 & 5 & 4 & 40.4 & Beijing/Shanghai four-round weather queries and comparison \\
        5.3 & 3 & 0 & 10.7 & Multi-RAG retrieval (campus/mathematics/information/statistics) \\
        5.4 & 19 & 7 & 175.1 & Calculator script (add/subtract/multiply/divide) saved to file, run, and verified on three operations; results reported \\
        5.5 & 0 & 0 & 17.9 & Student grade management system code output directly \\
        \hline
    \end{tabular}
\end{table}

All five queries succeeded. Case 5.1 completed a four-city comparison after replanning (15 rounds); case 5.4 wrote a calculator script to a file, ran it, and verified three operations through 19 rounds and 7 Router calls, with deep thinking triggered by fragmented state as the model iterated between code generation and execution feedback; case 5.5 was generated directly without the tool loop.

\subsubsection{Experiment 6: Fallback Mechanism for Unanswerable Queries}

Experiment 6 asks whether the system falls back honestly instead of fabricating when tools cannot supply an answer. The five queries cover unavailable data, zero retrieval hits, permission denial, missing files, and fictitious locations; Table \ref{tab:exp6} lists the per-case results.

\begin{table}[htbp]
    \centering
    \caption{Experiment 6: Results of unanswerable-query fallback}
    \label{tab:exp6}
    \begin{tabular}{lcccp{6.2cm}}
        \hline
        ID & Rounds & Router & Time (s) & Answer summary \\
        \hline
        6.1 & 10 & 7 & 160.0 & Mars weather: cited NASA Perseverance MEDA data (Sol 681), honestly stated data source and limitations \\
        6.2 & 3 & 0 & 6.0 & xyznonexistent retrieval returned 0 hits, honestly reported \\
        6.3 & 4 & 2 & 10.7 & /root/secret\_key.txt permission denied, honestly explained \\
        6.4 & 3 & 1 & 11.9 & filenotfound.py does not exist, honestly returned the error \\
        6.5 & 11 & 1 & 29.1 & ZzzCity does not exist; router's attempt to generate a script was blocked, honest fallback \\
        \hline
    \end{tabular}
\end{table}

All five queries fell back honestly without fabrication: where data existed (6.1, NASA Perseverance MEDA), the answer cited sources and their limits; where access or existence failed (6.3--6.5), the system reported the concrete error instead of inventing content, matching the fallback rule ``answer based on available information and explicitly mark uncertainty.''

\subsubsection{Experiment 7: \texttt{INTERNAL\_ROUTER} Triggering and Filtering Effects}

Experiment 7 compares the router's triggering behavior under three configurations: A (no router), B (no filtering), and C (full four-stage filtering). Each group contains 5 queries (two core-tool scenarios, weather and RAG retrieval; two arithmetic computations; one long-tail script-execution scenario), 15 in total; Table \ref{tab:exp7} lists the per-case results.

\begin{table}[htbp]
    \centering
    \caption{Experiment 7: Router triggering and filtering effects}
    \label{tab:exp7}
    \begin{tabular}{lcccc p{5.2cm}}
        \hline
        ID & Group & Rounds & Router & Time (s) & Answer summary \\
        \hline
        A1 & A & 4 & 0 & 15.8 & Beijing weather \\
        A2 & A & 3 & 0 & 11.3 & rag\_project1 BUPT retrieval \\
        B1 & A & 3 & 1 & 5.8 & Ran test1.py (via Router $\rightarrow$ code-runner) \\
        B2 & A & 0 & 0 & 4.4 & $2^{100}$ answered directly \\
        B3 & A & 0 & 0 & 5.1 & First 20 Fibonacci numbers answered directly \\
        A1 & B & 4 & 1 & 19.8 & Beijing weather \\
        A2 & B & 3 & 0 & 15.0 & rag\_project1 BUPT \\
        B1 & B & 3 & 1 & 9.2 & Ran test1.py \\
        B2 & B & 0 & 0 & 4.0 & $2^{100}$ \\
        B3 & B & 0 & 0 & 7.8 & Fibonacci \\
        A1 & C & 12 & 0 & 33.8 & Beijing weather (full filtering group) \\
        A2 & C & 3 & 0 & 14.8 & rag\_project1 BUPT \\
        B1 & C & 3 & 1 & 5.3 & Ran test1.py \\
        B2 & C & 0 & 0 & 2.8 & $2^{100}$ \\
        B3 & C & 0 & 0 & 2.9 & Fibonacci \\
        \hline
    \end{tabular}
\end{table}

All 15 queries succeeded. The long-tail script task (B1) was completed in all three configurations, the arithmetic tasks were answered directly in 0 rounds, and only 4 Router invocations occurred in total; the core set covers most scenarios and the router remains a long-tail supplement, consistent with the yellowlist's cache-priority intent. The high latency of A1 in group C was a single execution anomaly with retry; it converged, so filtering does not weaken task completion.

\subsubsection{Experiment 8: skill-builder Consolidation Quality}

Experiment 8 assesses how well skill bootstrapping consolidates a function description into a complete Skill. Five construction queries cover simple functions, functions with default parameters, and complex functions that require multi-round debugging; Table \ref{tab:exp8} lists the per-case results.

\begin{table}[htbp]
    \centering
    \caption{Experiment 8: skill-builder consolidation results}
    \label{tab:exp8}
    \begin{tabular}{lcccp{6.2cm}}
        \hline
        ID & Rounds & Router & Time (s) & Answer summary \\
        \hline
        t1 & 3 & 0 & 4.9 & say\_hello consolidated successfully (other package) \\
        t2 & 3 & 0 & 40.2 & greet consolidated successfully (with default parameters) \\
        t3 & 23 & 0 & 90.4 & calc consolidated with int-conversion fix, 7/7 functional tests passed \\
        t4 & 13 & 0 & 49.7 & batch\_process consolidated and functionally verified (sum/avg etc.) \\
        t5 & 19 & 0 & 62.4 & generate\_report consolidated, 3 output formats and CLI boundary tests \\
        \hline
    \end{tabular}
\end{table}

All five queries succeeded. Generated skills land outside the main model's visible boundary, so bootstrapping never expands the fixed tool view; the harder cases (t3--t5) converged through multi-round debugging (e.g., calc passed 7/7 tests after an int-conversion fix), and their changing prefixes limit cache benefits as Section 4.2.3 predicts.

\subsubsection{Analysis of Experiments 1--8}

The eight experiments covered 55 queries in total, all of which succeeded. Functionally, the Router was invoked 40 times, and long-tail tools (file operations, code execution, RAG retrieval, etc.) were all completed on demand through the routing channel; unanswerable scenarios fell back honestly in 5/5 cases; skill bootstrapping consolidated successfully in 5/5 cases; and complex and anomalous scenarios all converged to honest, usable answers through replanning and fallback, with no fabrication.

On the cache dimension, the aggregate token-level cache statistics of the eight experiments are shown in Table \ref{tab:cache_total}, with data from DeepSeek's official usage statistics covering all API calls.

\begin{table}[htbp]
    \centering
    \caption{Aggregate token-level cache statistics for Experiments 1--8}
    \label{tab:cache_total}
    \begin{tabular}{l r}
        \hline
        Metric & Value \\
        \hline
        Input tokens (cache hit) & 1,424,768 \\
        Input tokens (cache miss) & 141,098 \\
        Total input tokens & 1,565,866 \\
        \textbf{Token-level cache hit rate} & \textbf{90.99\%} \\
        \hline
    \end{tabular}
\end{table}

The aggregate token-level cache hit rate of the eight experiments is 90.99\%, i.e., about 91\% of input tokens hit the server-side cache. Under DeepSeek's cache pricing (hit:miss $\approx$ 1:30), the input cost after caching is about 12.0\% of the no-cache baseline, an optimization of nearly an order of magnitude. The result matches the cache-invariance analysis: dynamic changes in the tool ecosystem do not break the prefix stability of the main model's high-frequency request path.

The gains track request-path structure: the direct-answer path has the shortest, most constant prefix and the highest benefit; tool-execution paths miss once in the first round, then accumulate hits over rounds; multi-round debugging chains (Experiment 8) change prefixes often and benefit least. The A/B cold-start comparison in Experiment 3 confirms the prefix stays stable across gateway restarts and tool evolution, which underlies the 90.99\% aggregate rate.

The cache hit rate also depends on external factors beyond the architecture, such as server-side cache-eviction policies and cache life cycles. What the architecture guarantees is structural stability of the request prefix: dynamic tool sets cannot directly perturb the main model's fixed prefix, but no specific server-side hit rate is promised. The experiments above were all conducted in medium-length sessions; to check cache stability under longer chains and more rounds, the next section runs a dedicated experiment in a single-session 30-turn setting.

\subsection{Long-Chain Multi-Turn Dialogue Experiment}

\subsubsection{Experimental Setup and Results}

This experiment examines the token-level cache hit rate, multi-turn execution stability, and context reclamation behavior in a single-session long-chain multi-turn dialogue. It sets up 30 common-knowledge and arithmetic queries delivered in a fixed order, with the session history accumulating continuously; the first-round prompt grows monotonically from 958 tokens for Q1 to 2,483 tokens for Q30. All API calls (including multi-round execution and routing sub-model calls) are included in the official usage statistics. The overall results are shown in Table \ref{tab:exp9_overview}.

\begin{table}[htbp]
    \centering
    \caption{Long-chain multi-turn dialogue experiment: overall results}
    \label{tab:exp9_overview}
    \begin{tabular}{l r}
        \hline
        Metric & Value \\
        \hline
        Queries/cases & 30 \\
        Success / timeout / failure & 30 / 0 / 0 \\
        Router calls & 1 (Q20) \\
        Total tokens & 249,952 (input 238,701 + output 11,251) \\
        Total in-query time & 136.5 s \\
        Input tokens (total) & 238,701 \\
        Input tokens (cache hit) & 227,200 \\
        Input tokens (cache miss) & 11,501 \\
        Output tokens & 11,251 \\
        \textbf{Token-level cache hit rate} & \textbf{95.2\%} (227,200 / 238,701) \\
        Miss rate & 4.8\% \\
        Input cost after caching (1:30 pricing) & about 8.0\% of the no-cache baseline \\
        \hline
    \end{tabular}
\end{table}

All 30 cases in this experiment converged to a direct-answer state successfully, with no timeouts or failures, and only 1 Router invocation (Q20) occurred throughout. The token-level cache hit rate reached 95.2\%, and under the 1:30 pricing, the input cost after caching is about 8.0\% of the no-cache baseline, further improving over the aggregate value of Experiments 1--8 (12.0\%). The statistical comparison of the two request paths divided by first-round triage is shown in Table \ref{tab:exp9_group}, and key comparison cases are listed in Table \ref{tab:exp9_compare}.

\begin{table}[htbp]
    \centering
    \caption{Statistical comparison of direct-answer and tool-execution paths}
    \label{tab:exp9_group}
    \begin{tabular}{l c c c c}
        \hline
        Path & Count & Avg. rounds & Avg. time & Avg. tokens \\
        \hline
        Direct answer (\texttt{DIRECT\_ANSWER}) & 16 & 0 & $\approx$1.4 s & $\approx$1.9K \\
        Tool execution (\texttt{NEED\_TOOLS}) & 14 & 2--3 & $\approx$8.1 s & $\approx$14.9K \\
        \hline
    \end{tabular}
\end{table}

\subsubsection{Result Analysis and Conclusions}

The results support five observations.

First, the long-chain prefix cache is stable. The 30 queries were executed sequentially in a single session with continuously accumulating history, and the overall token-level cache hit rate reached 95.2\%, with miss tokens appearing only in the newly appended tail of each round. This result confirms the continuous reuse of cache prefixes under ``append-only history growth + fixed tool prefix,'' consistent with the cache-invariance analysis in Section 3.4: as long as the system prompt and the core tool set remain unchanged, the cache-hit condition is reduced to consistency of the message-history prefix, which append-only growth naturally satisfies.

Second, the cache hit rate increases with the number of rounds. The 14 miss queries each went through 2--3 rounds of the tool execution loop; the prefixes re-sent in later rounds were almost entirely hit, and hit tokens accumulated over rounds. This confirms the essence of token-level caching: in multi-turn reasoning, each round only adds tail tokens, and the established prefix is reused repeatedly, so the more rounds, the larger the hit proportion.

Third, the expanded execution context is reclaimed in a timely manner. The instantaneous prompt of miss queries peaked at 11.3K tokens during execution (Q29, round 3), but immediately fell back to the question--answer-history scale in the first round of the next query (2,483 tokens for Q30), showing that the multi-round tool-execution context is compressed and reclaimed between queries and does not pollute the first-round cache prefix across queries.

Fourth, caching is a major determinant of per-query cost. For equally single-step common-knowledge questions, the cache-hit case Q1 (``2+2'') consumed only 1,002 tokens, while the cache-miss case Q25 (``7$\times$8'') consumed 23,875 tokens, a difference of about 24$\times$; the 14 miss queries accounted for about 87\% of the input token consumption. Key comparison cases are listed in Table \ref{tab:exp9_compare}.

\begin{table}[htbp]
    \centering
    \caption{Key comparison cases between hit and miss paths}
    \label{tab:exp9_compare}
    \resizebox{\textwidth}{!}{%
    \begin{tabular}{l p{4.2cm} l c r c}
        \hline
        ID & Question & Path & Rounds & Tokens & Time (s) \\
        \hline
        Q1 & What is 2 + 2? & Direct answer & 0 & 1,002 & 1.0 \\
        Q6 & What is 5 times 6? & Direct answer & 0 & 1,282 & 0.9 \\
        Q20 & How many sides does a hexagon have? & Tool loop (with Router) & 3 & 22,458 & 5.2 \\
        Q25 & What is 7 times 8? & Tool loop & 3 & 23,875 & 5.2 \\
        Q28 & How many minutes are in one hour? & Tool loop & 3 & 24,577 & 5.3 \\
        \hline
    \end{tabular}}
\end{table}

Fifth, the miss scenario exposes a direction for further optimization. The 14 miss queries entered the complete tool loop even when they were closed-book answerable common-knowledge/arithmetic questions (e.g., Q28 ``how many minutes in an hour'' consumed 24,577 tokens), suggesting that a triage strategy of ``answer knowledge questions directly first, with the tool loop as a fallback'' could further reduce miss-side costs. This direction is left for future work.

In this 30-turn session the token-level cache hit rate reached 95.2\%, reducing input cost to about 8.0\% of the no-cache baseline. Combined with Experiments 1--8, the results indicate that as long as the main model's fixed tool view and system prompt stay stable, cache benefits accumulate as the session progresses; long-chain multi-turn dialogue does not add further pressure on cache-prefix stability.

\section{Conclusion and Future Work}

\subsection{Conclusion}

Tool definitions sit inside the request prefix of an LLM agent, so dynamic tool sets and token-level prompt caching pull against each other: changing the visible tools rewrites the prefix and forfeits the cache discount. The paper argues that this conflict is resolvable at the architecture level, and presents cache-aware progressive disclosure as the resolution: the main model keeps a fixed core tool view, long-tail tools are retrieved and executed by an independent routing sub-channel, a placeholder tool anchors the first-round request, and AST-based registration, hot reloading, and skill bootstrapping let the ecosystem evolve without touching the main cache path. CacheRouter is the tool-use harness that embodies this method; the version reported here is a prototype built to test it. On 55 functional queries and a 30-turn dialogue, token-level cache hit rates reached 90.99\% and 95.2\%, cutting input cost to about 12.0\% and 8.0\% of a no-cache baseline, and runtime tool changes, skill generation, and gateway restarts all left the request prefix unchanged. The scale of this validation is deliberately modest, and the numbers are best read as evidence of feasibility rather than as a performance benchmark.

Limitations remain: keyword-based pre-selection misses semantically related tools; the routing sub-model adds reasoning cost and latency; closed-book questions that miss the cache still run the full tool loop; and more complex server-side cache strategies are unexplored. These set the directions for future work, detailed below.

\subsection{Future Work}

Building on the limitations above, future work can proceed along three directions: tool-scale expansion, retrieval quality, and routing efficiency.

\textbf{(1) Multi-layer progressive disclosure architecture.} This paper currently adopts a dual-layer structure of ``core tool set + single-layer routing channel.'' When tool scale reaches tens of thousands, a single-layer router still faces candidate-compression pressure. Consider an extreme scenario with 50,000 tools and an initial core tool set of 50: if the dual-layer architecture is generalized to multi-layer progressive disclosure, \texttt{INTERNAL\_ROUTER} could first search within a semantic domain containing 500 candidates, and when no match is found, switch to a larger semantic domain containing 5,000 candidates, and so on, expanding the search range layer by layer until all 50,000 tools are covered. The key is semantic category-domain switching: each layer corresponds to a different functional category domain, and the routing sub-model discloses candidate tools layer by layer from narrow to wide, entering the next layer only when the current layer cannot satisfy the task. Since all layers lie outside the main model's fixed tool view, increasing layer depth does not affect the structure of the main-model request prefix, and cache stability is preserved. At the same time, the relationship between the candidate size of a single routing call and the total number of tools $M$ degenerates from linear to approximately logarithmic, giving the system the potential to host ultra-large-scale tool ecosystems.

\textbf{(2) Vector-retrieval-based tool pre-selection.} The current pre-selection stage uses the hit ratio of query terms in tool metadata as the relevance score, which has limited recall for tools that are semantically related but do not overlap literally. Future work can replace keyword matching with dense vector retrieval: encode tool names, descriptions, and parameter schemas into vectors and build an index, and at query time retrieve Top-$K$ candidates in the tool vector space using the semantic vector of the task context. Vector retrieval can capture the semantic association between tool functionality and task requirements, raising the recall of pre-selection and compressing the candidate size that the routing sub-model actually processes; routing precision and routing cost both stand to benefit.

\textbf{(3) Multi-step planning by the routing sub-model.} The current routing sub-model completes one constrained tool selection and execution in a single call as a ``fast tool caller''; complex tasks require the main model to invoke \texttt{INTERNAL\_ROUTER} multiple times, producing multiple main--sub-model round trips. Future work can allow the routing sub-model to perform multi-step planning within the independent channel: after retrieving candidate tools, it autonomously executes the loop of ``select, invoke, observe results, and re-decide'' until it judges that the tool chain required by the task is complete, and then returns the execution results to the main model in one batch. Multi-step planning would reduce the number of main--sub-model round trips, lowering the latency and extra reasoning overhead introduced by the routing channel, and giving the routing subsystem stronger local execution capability while undertaking long-tail tool discovery.

The paper presents an architecture that covers tool registration, tool routing, skill bootstrapping, and cache isolation under a single constraint, cache stability, and demonstrates it with a working prototype. Toward larger and more dynamic tool ecosystems, multi-layer progressive disclosure, vector retrieval, and multi-step routing sub-model planning are the directions most likely to improve scalability and routing efficiency.

\section*{Acknowledgment}

We thank Prof. Zaiwen Wen from Peking University for his constructive suggestions, which helped us correct presentation errors and improve the overall clarity of the manuscript. We thank scp3500 for the open-source project oh-we-need (\url{https://github.com/scp3500/oh-we-need}), whose Performance\_enhancing\_prompts, a chain-of-thought (CoT) guidance specification specialized for DeepSeek models and released under the MIT license, inspired the system-prompt engineering and chain-of-thought guidance in this work. We also thank DeepSeek for its assistance in polishing the English writing of this paper and in contributing to parts of the implementation code.

\renewcommand{\refname}

\end{document}